\documentclass{article}
\usepackage{spconf,amsmath,graphicx}
\usepackage[symbol]{footmisc}
\usepackage{subcaption}
\usepackage{hyperref}
\usepackage{tabularx}
\usepackage{multirow}
\usepackage{booktabs}
\usepackage{xurl}
\usepackage{siunitx}

\title{ADAPTING THE ADAPTERS FOR CODE-SWITCHING IN MULTILINGUAL ASR}
\name{Atharva Kulkarni$^{*,\ddagger}$, Ajinkya Kulkarni$^*$, Miguel Couceiro$^{\dagger}$, Hanan Aldarmaki$^*$}
\address{$^*$ MBZUAI, UAE, $^{\dagger}$ Université de Lorraine, CNRS, LORIA, France, $^{\ddagger}$Erisha Labs, India}
\begin{document}
%
\maketitle
\begin{abstract}


Recently, large pre-trained multilingual speech models have shown potential in scaling Automatic Speech Recognition (ASR) to many low-resource languages. Some of these models employ language adapters in their formulation, which helps to improve monolingual performance and avoids some of the drawbacks of multi-lingual modeling on resource-rich languages. However, this formulation restricts the usability of these models on code-switched speech, where two languages are mixed together in the same utterance. In this work, we propose ways to effectively fine-tune such models on code-switched speech, by assimilating information from both language adapters at each language adaptation point in the network. We also model code-switching as a sequence of latent binary sequences that can be used to guide the flow of information from each language adapter at the frame level. The proposed approaches are evaluated on three code-switched datasets encompassing Arabic, Mandarin, and Hindi languages paired with English, showing consistent improvements in code-switching performance with at least 10\% absolute reduction in CER across all test sets. 



\end{abstract}
\section{Introduction}
\label{sec:intro}

Large pre-trained multi-lingual speech models have been proliferating in recent years, including various self-supervised models, as well as supervised ASR models, such as the recently released Massively Multilingual Speech (MMS) project\footnote{\url{https://ai.meta.com/blog/multilingual-model-speech-recognition/}}, which expanded the multilingualism in speech processing applications to 1,406 languages.
To account for the variability in these languages, language adapters were employed to learn language-specific features at some parts of the networks, while sharing most of the overall parameters to enable cross-lingual transfer. While this formulation leads to better management of the vast multi-lingual use cases, it restricts the use of the model on code-switched speech, 
where a single speech utterance contains two or more languages. Code-switching (CS) is generally challenging for ASR models, 
which usually result in degraded ASR performances \cite{csfuture}. This decline in performance in ASR tasks while handling CS may lead to error propagation across various modules (both text and speech processing) in conversation AI systems. 
CS  can be categorized into two types: inter-sentential CS, where language switches occurs at sentence boundaries, and intra-sentential CS, where the switching takes place within the same sentence \cite{litrew1}. The \textit{matrix} language is the main language used as a basis in most utterances, and the \textit{embedded} language is the secondary language in code-switching. For example, many Arabic or Hindi speakers often use English as an embedded language in casual conversations.

Previous works that attempt to address CS in bilingual or multilingual ASR \cite{litrew2,litrew3,litrew4} typically work by explicitly modeling language information either in acoustic ASR models \cite{w2vecindic,ctclmw2vec,csafr,csasrlang} or in language modeling \cite{litrew5,litrew6,lmcsasr}. 
In contrast to previous works that train models for specific language pairs from scratch, we focus on improving CS recognition in large pre-trained multilingual models that rely on language adapters, using MMS as our model of choice. MMS is based on the Wav2Vec 2.0 architecture and has been trained both using self-supervised learning as well as supervised pre-training for ASR using language adapters that can be loaded and exchanged at inference time as needed. The model consists of multiple Transformer blocks, each appended with a language-specific adapter, in addition to language-specific softmax layers for prediction. 


\begin{figure*}[!t]
  \centering
 \includegraphics[scale=0.77]{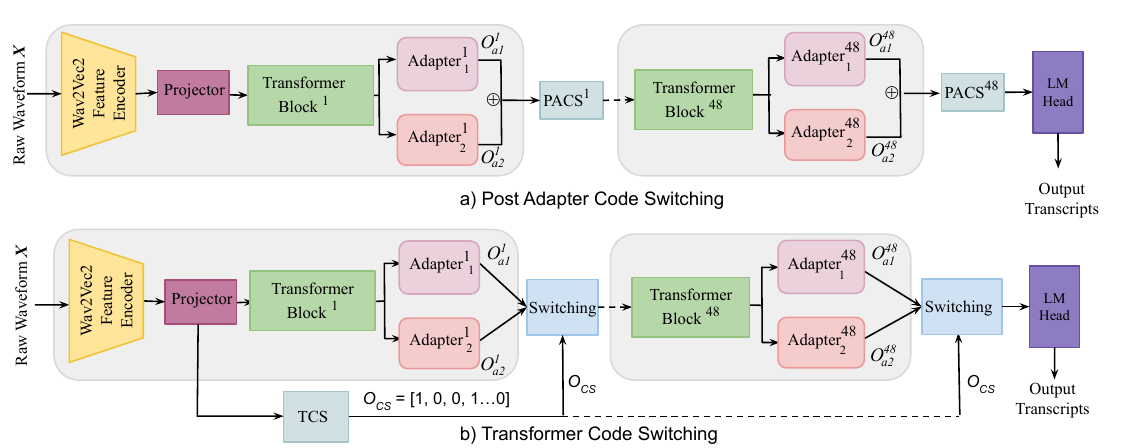}
    \caption{Framework of proposed approaches with MMS, \textbf{a) Post Adapter Switching Approach b) Transformer Code Switching.} Transformer blocks range from 1 to 48 and the grey color indicates frozen model parameters during training.}
    \label{fig:main}
   \vspace{-0.8em}
\end{figure*}



As intra-sentential CS speech contains more than one language within utterances, the pre-specification of a language adapter results in degradation of ASR performance since the model is not optimized to identify and predict the embedded language. In this work, we introduce two approaches to regulate the information flow through each transformer block using two language adapters. First, we propose a {\it Post-Adapter Switcher} (PAS) network, which integrates the information from two adapters and learns to modulate the hidden output information. Second, we propose a {\it Transformer Code Switcher} (TCS), where we explicitly estimate the binarized sequence of CS points using a transformer block, which enable adapter switching at the frame level. The latent binary code sequence is assimilated in the hidden outputs from both target language adapters in each block of transformer in ASR, thus enabling the MMS system to gain finer control over boundaries of CS in a given speech utterance. The PAS and TCS network are first fine-tuned using relatively small amounts of CS training data in the target language pair, while keeping all other parameters frozen.
To illustrate the robustness of the approach, we evaluate it on three different CS datasets, for which English is the embedded language: ASCEND (Mandarin-English) \cite{ascend}, ESCWA (Arabic-English) \cite{lm_csasr}, and MUCS (Hindi-English) \cite{mucs}. We released the pre-trained models along with inference code on a Github repository\footnote{{\url{https://github.com/Atharva7K/MMS-Code-Switching}}}.



\noindent The \textbf{ contributions} of this paper are as follows, \textbf{1.} Presenting a novel adapter switching network for handling CS in MMS and similar systems. \textbf{2.} 
Proposing 
a novel mechanism for regulating information flow from language-specific adapters at the frame level. \textbf{3.} Demonstrating improved performance on CS datasets compared to the original MMS model.

\section{Proposed Approaches}
\label{sec:prop_approach}

In this section, we present the proposed framework for adapting the adapters in the context of code-switched ASR using the MMS system. First, we briefly describe the general framework of MMS ASR system and the role of adapters in scaling it to multiple languages. Thereafter, we provide the details of the proposed approaches to utilize two language adapters for recognizing CS speech: {\it Post Adapter Code Switching} (PACS) and {\it Transformer Code Switching} (TCS). 



\subsection{The MMS model}

The MMS framework is built upon the Wav2Vec2 architecture, which serves as the feature encoder network. This encoder learns to convert raw speech waveforms into latent representations using 1D temporal convolutions. In the ASR task, these latent representations are input into transformer blocks to generate contextual representations. The final contextual representations are processed by a language-specific feed-forward/softmax layer, called LM head, to predict the distribution of tokens within the language vocabulary. The transformer block follows the BERT architecture \cite{bert} with 48 blocks. MMS training involves a two-stage process. First, it undergoes self-supervised learning by solving a contrastive task using masked feature encoder outputs from training data representing 1,406 languages. After this pre-training phase, the base model weights, including the feature encoder, feature projector, and the 48 transformer blocks, are frozen. Only the adapter modules and the LM head are further trained on labeled data using the CTC loss for ASR. During inference, MMS only requires specification of the language as an additional argument along with the speech utterance to load the corresponding language adapters and LM head. For more comprehensive details on MMS and the ASR training setup, please refer to \cite{mms}.

To support ASR for over 1,100 languages, language-specific adapter modules \cite{adapters} are incorporated at the end of each transformer block. These adapter modules consist of two feedforward layers with a LayerNorm layer and a linear projection to 16 dimensions with ReLU activation. Additionally, language-specific LM heads are selectively trained because token characters can vary among languages. This selective approach is especially important in cases like CS ASR, where the system needs to produce token characters from multiple languages, making a single language-specific adapter inadequate.

\subsection{Post Adapter Code Switching (PACS)}

This section outlines the Post Adapter Code Switching (PACS) approach, which involves incorporating information from two pre-trained language adapters, denoted as $A_{1}^{n}$ and $A_{2}^{n}$ corresponding to the matrix and embedded languages, where subscript $n$ denotes the $n^{th}$ transformer block. 
As depicted in Fig \ref{fig:main}a, the hidden output of each transformer block is fed into both $A_1^{n}$ and $A_2^{n}$. The resulting outputs, $O_{a1}^n$ and $O_{a2}^n$ from $A_{1}^n$ and $A_{2}^n$, are both 1280-dimensional. These outputs are concatenated and provided as input to the $n^{th}$ PACS block, which is designed with the same architecture as the adapters and has an input dimension of 2560, allowing it to accommodate the concatenated outputs from $O_{a1}^n$ and $O_{a2}^n$. 
During the training phase, MMS is trained alongside PACS on a labeled code-switching (CS) speech dataset. This training inherently enables PACS to learn how to handle CS without the need for additional auxiliary loss. The weights of the Wav2Vec2 encoder and transformer blocks remain fixed.  The output dimension of the PACS module is set to 1280 to fit the input dimension for the subsequent transformer block. This design facilitates the seamless integration of PACS into the MMS framework without modifying the underlying MMS components.

Furthermore, to consider character tokens from both languages in CS, we concatenate the pretrained weights of the LM head layers from both languages. For example, if the English and Arabic LM Head layers have output dimensions of 154 and 121 respectively, we created a new feedforward layer with a dimension of 275 (154+121). We then copy the first 154 weights from the English LM head and the remaining 121 from the Arabic LM head. This approach allows us to utilize pre-trained  LM head without the need to train them from scratch. Additionally, it's worth noting that each LM head contains lower-case English alphabets along with punctuation and digits. This presents a challenge when merging the pre-trained LM head weights of the English and matrix languages. To address this, we construct a binary mask that assigns a probability of $0.0$ to punctuation and English alphabet character tokens for the matrix language LM head. The concatenated LM head is fine-tuned along with the code-switching network.  

\subsection{Transformer Code Switching (TCS)}

In this section, we describe Transformer Code Switcher (TCS), which is designed to capture the latent binary sequence representing CS points in the utterance. The TCS network takes the output of the projector block as input to estimate switching points from input features, 
as depicted in Fig \ref{fig:main}b. The estimated binary code sequence, denoted as $O_{CS} = [1, 1, ..., 0, 1, ... 0]$, is combined with the outputs of the two adapters, namely $O_{a1}$ and $O_{a2}$, at each transformer block, according to Equation 1. 
\begin{equation}
O_{Switch}^{n} =  (1-O_{CS}) *  O_{a1}^{n} + O_{CS} * O_{a2}^{n}
\end{equation} 

For implementing TCS, we chose a transformer architecture with an output sigmoid activation function. This choice allows us to leverage the self-attention mechanism, which assists in identifying CS boundaries at the frame level. Therefore $O_{CS}$ regulates the flow of information from two adapters to enable the network to handle CS speech by dynamically masking out one of the languages based on the input features.  We applied a threshold value of 0.5 to the output of the sigmoid activation to create binarized latent codes for CS. Subsequently, 
the outputs of both adapters are processed through Equation 1 as $O_{Switch}^n$ and passed to the next transformer block. Thus, the switching operation enables finer control over CS. Since the output dimension of adapters is of 1280 dimension, we repeated $O_{CS}$ across the output dimension to ensure the mask is applied across all features in the given frame.

\section{Data Preparation}
\label{sec:data_prep}


		 

We use 3 code-switched datasets to evaluate our models. \textbf{ASCEND} \cite{ascend}: is a Mandarin-English code-switched corpus with spontaneous conversational speech spanning over 10.62 hours comprising of \~12.3K utterances in total. The train-validation-test splits contain 9.8K, 1.1K, and 1.3K utterances corresponding to 8.78, 0.92, and 0.92 hours of speech, respectively. \textbf{MUCS} \cite{mucs}: is a family of multilingual and code-switched ASR datasets for Indic languages introduced in the MUCS challenge. We use only the Hindi-English code-switched data from MUCS with a train-val split of 52.8K, 3.1K utterances and duration 89.86, and 5.18 hours of speech, respectively. For evaluation, we use the validation set as the test set as no explicit evaluation set was provided.
\textbf{QASR} \cite{qasr}: is a large-scale multi-dialectal speech corpus for Arabic spanning over 2000 hours with some utterances containing Arabic-English CS. We selectively use only the code-switched samples from the entire corpus for training. Specifically, we select transcripts that contain complete English words and not only numbers or dates. We were able to extract 5.8K utterances over 4.8 hours for the training. \textbf{ESCWA}: \cite{lm_csasr} It consists of 845 utterances spanning 2.8 hours with Arabic-English code-switching. We used ESCWA as the test set to evaluate models trained on the QASR dataset.

\label{sec:results}

\begin{table*}[]
\centering
\small
 \begin{tabular}{|l|ll|ll|ll|l|}
    \hline
    \multirow{2}{*}{Model} & \multicolumn{2}{l|}{ASCEND} & \multicolumn{2}{l|}{ESCWA} & \multicolumn{2}{l|}{MUCS}  & \multicolumn{1}{l|}{\# params}  \\ 
    \cline{2-7}
    & {MER} & {CER} & {WER} & {CER} & {WER} & {CER} & {(Millions)} \\ \hline
     \multicolumn{8}{|l|}{\textbf{MMS with single language adapter:}} \\ \hline
    English             & 98.02  & 87.85 & 92.73 & 71.14 & 101.72 & 74.02 & 965  \\
    Matrix-language            & 71.98  & 66.76 & 75.98 & 46.38 & 58.05  & 49.20 & 965  \\ \hline
    \multicolumn{8}{|l|}{\textbf{Proposed models for fine-tuning:}} \\ \hline
    Matrix-language-FT   & 45.97  & 44.13 & 77.47 & 37.69 & 66.19  & 41.10 & 965 \\
    PACS                & 44.41  & 40.24 & 75.50 & 46.69 & 63.32  & 42.66 & 970  \\ 
    TCS                 & 41.07  & 37.89 & 74.42 & \textbf{35.54} & \textbf{57.95}  & \textbf{38.26} & 980 \\ \hline
    \multicolumn{8}{|l|}{\textbf{Whisper-large, with language id prompt:}} \\ \hline
      Matrix language id 
      & 23.31  & 23.49 & 59.30 & 39.66 & 104.4 & 87.48 & 1550   \\
     Concatenated prompt \cite{peng2023prompting}      &  \textbf{21.13} & \textbf{21.30}  & \textbf{58.70} & 38.86 & 101.62  & 91.59  & 1550  \\ \hline 
  \end{tabular}
\caption{Evaluation of proposed approaches on three CS datasets (ASCEND, ECSWA, and MUCS).
Except for Whisper and Whisper-prompt, all other systems indicate the adapter approach with MMS; FT denotes fine-tuned.}
\label{tab:mytable}
\end{table*}

\section{Experimental setup}
\label{sec:exp_setup}
All the experiments are conducted by modifying the HuggingFace implementation\footnote{\href{}{\url{https://github.com/huggingface/transformers/tree/main/src/transformers/models/wav2vec2}}}. 
We used Adam optimizer with a learning rate of 1e-6 in the case of QASR, ASCEND, and 1e-5 for MUCS to optimize our models using CTC loss. We experimented with learning rates of 1e-3, 1e-5, and 1e-6 that led to best performances. We used a warm-up of 1K steps and decaying thereafter according to a polynomial lr scheduling. Following the experimentation with batch sizes of 8, 16, and 32, we found that a batch size of 32 is optimal for QASR and ASCEND, while a batch size of 16 is most effective for MUCS. All the experiments were carried out on a single Nvidia A100 40GB GPU for up to 100 epochs, by selecting the best-performing model on the validation set where applicable. For the CTC loss function, we set the \texttt{ctc\_zero\_infinity} parameter to True for managing optimization issues. 
\vspace{-1em}
\section{Results and Discussion}

We report performance in terms of ASR character error rate (CER) and mixed error rate (MER) to evaluate performance on Mandarin-English code-switched ASCEND, which accounts for the fact that individual Chinese characters represent words. To compute CER and MER for ASCEND, we used the implementation from the official  
 repository \footnote{\href{ASCEND}{\url{https://github.com/HLTCHKUST/ASCEND/blob/main/eval.py}}}.In case of ESCWA and MUCS datasets, we use Word Error Rate (WER) and CER for evaluation using HuggingFace evaluate library\footnote{\url{https://huggingface.co/docs/evaluate}}. 

Table \ref{tab:mytable} shows the performance of all of our approaches along with the baselines on the three datasets. We can see that plain MMS performs poorly on code-switched datasets. In the case of plain MMS, the performance is better when we use language adapters corresponding to the matrix language, {\it i.e}, Mandarin in case of ASCEND, Hindi in the case of MUCS,  and Arabic in the case of ESCWA/QASR as expected. 
We also directly fine-tune the matrix language adapter as another baseline, termed Matrix-language-FT in the Table 1. 
Finetuning matrix-language adapters significantly improves the CER in case of ESCWA and MUCS, but it worsens the WER. 
Except  ASCEND, fine-tuning actually degrades WER; this indicates that directly fine-tuning one language adapter on CS speech is not a feasible option, as the pre-trained adapters are largely biased towards the specified languages. 
Both proposed frameworks, PACS and TCS improve performance compared with the original model and direct fine-tuning of the matrix language adapter.  Among MMS-based variants, TCS results in the best overall performance in all metrics and test sets. 

We also show the performance of the pre-trained Whisper-large model as a point of reference since the two models are not directly comparable. We do not fine-tune the Whisper model, but we use the prompt approach described in \cite{peng2023prompting}, which concatenates the language ids in the prompt  to improve performance on CS speech.  One interesting observation from these results is that Whisper model, without fine-tuning, results in much better performance than MMS raw performance (over 40\% absolute CER difference) in the ASCEND dataset.  While our best adaptation method improves performance by $\sim$30\% absolute CER, it remains significantly higher the Whisper's raw performance. This is likely a direct result of the size of the labeled dataset used in the supervised pre-training. While exact numbers of training hours are not directly reported for MMS, we can estimate it based on the component datasets. In our estimate, MMS ASR model was pre-trained on $\sim$1K hours of Mandarin speech, while Whisper was pre-trained using over 23K hours. 



\section{Conclusion}
\label{ssec:conclusion}

We presented a framework for adapting adapter-based multi-lingual ASR models for code-switched speech recognition. We conducted experiments using the MMS model, which was pre-trained on a wide range of languages, including English, Mandarin, Arabic, and Hindi, matching the languages in our CS test sets. Since the MMS framework involves loading language adapters, the original model performs poorly in our test sets. The proposed models involve loading the language adapters for the matrix and embedded language, and mixing their predictions to produce better code-switched transcriptions. With a small increase in model size, our proposed method improves the CS output across all test sets, with more than 10\% absolute reduction in CER. Note however that we did not employ language model rescoring in the current empirical study. Thus our models could be potentially improved by using an external LM trained with synthetic CS text or by controlling the adapter choice by reinforcement learning, which we leave for future work.  



\newpage
\bibliographystyle{IEEEbib}
\bibliography{Template}

\end{document}